# Nurse-in-the-Loop Artificial Intelligence for Precision Management of Type 2 Diabetes in a Clinical Trial Utilizing Transfer-Learned Predictive Digital Twin

Syed Hasib Akhter Faruqui[1], Adel Alaeddini[2], Yan Du[3], Shiyu Li[3], Kumar Sharma[3], Jing Wang[4]

[1] Sam Houston State University, Huntsville, TX
[2] The University of Texas at San Antonio, San Antonio, TX
[3] The University of Texas Health San Antonio, San Antonio, TX
[4] Florida State University, Tallahassee, FL

## Abstract:

**Background:** Type 2 diabetes (T2D) is a prevalent chronic disease with a significant risk of serious health complications and negative impacts on the quality of life. Given the impact of individual characteristics and lifestyle on the treatment plan and patients' outcomes, it is crucial to develop precise and personalized management strategies. Artificial intelligence (AI) provides great promise in combining patterns from various data sources with nurses' expertise to achieve optimal care.

**Methods:** This is a 6-month ancillary study among T2D patients (n = 20, age = 57 $\pm$ 10). Participants were randomly assigned to an intervention (AI, n=10) group to receive daily AI generated individualized feedback or a control group without receiving the daily feedback (non-AI, n=10) in the last three months. The study developed an online nurse-in-the-loop predictive control (ONLC) model that utilizes a predictive digital twin (PDT). The PDT was developed using a transfer-learning based Artificial Neural Network. The PDT was trained on participants' self-monitoring data (weight, food logs, physical activity, glucose) from the first three months, and the online control algorithm applied particle swarm optimization to identify impactful behavioral changes for maintaining the patient's glucose and weight levels for the next three months. The ONLC provided the intervention group with individualized feedback and recommendations via text messages. The PDT was re-trained weekly to improve its performance.

**Findings:** The trained ONLC model achieved ≥80% prediction accuracy across all patients while the model was tuned online. Participants in the intervention group exhibited a trend of improved daily steps and stable or improved total caloric and total carb intake as recommended. The intervention group exhibited significant weight loss (p-value 4.731e-8, average:5.871 lbs.) and maintained glucose levels over time (p-value 0.661) compared to baseline.

**Interpretation:** Nurse-in-the-loop AI using a digital twin approach can potentially improve adherence to recommended healthy lifestyle behaviors. Further studies with larger sample sizes and long-term follow-ups are warranted.

# Introduction:

Recently, there has been a rise in the implementation of machine learning algorithms to improve blood glucose control [14]. Machine learning models that have been implemented so far include support vector machines [1], gaussian process [2], ensemble methods [3], multilayer perceptron [4], time series convolution neural networks [5], deep convolutional neural networks [6], and RNN & LSTM models [5,7], etc. Zhu et al.[7] systematically reviewed deep learning algorithms for diabetes diagnosis, complication diagnosis, and glucose management. In addition, researchers have proposed using transfer learning strategies for small and imbalanced datasets, which also gives them the advantage of fine-tuning for specific patients [8,9]. This also gives the researchers the advantage of fine-tuning for specific patients [8,9]. Faruqui et al.[9] developed similar algorithms that utilize a transfer-learning methodology to fine-tune models for patient-specific characteristics, which focuses on how the amount of macro-nutrients (fat, carb, fiber, protein, physical activity, and previous days' blood glucose levels) affects the patient blood glucose level and weight.

To mitigate extended delays, several optimization and controller-based models have been developed to automate the insulin delivery system [10–12]. Most of these studies focus on controlling insulin infusion devices [12,13] to control blood glucose levels. However, only a few studies cover the effects of macro-nutrients on a patient's blood glucose level and weight [14]. Furthermore, various factors determine blood glucose levels, including how much fat, carb, fiber, and protein patients consume and how much physical activity they have engaged in [14]. Again, most studies focus on controlling insulin infusion devices rather than providing suggestions to patients to support their pursuit of self-control in their diet and activity behaviors, which are key to weight and glucose control in T2D patients. Thus, there is a pressing demand for highly efficient online models capable of considering a patient's real-time condition and subsequently generating supportive measures, such as motivational messages, dietary recommendations, and physical activity guidance, among others.

This study leverages a transfer-learned digital twin model to develop a nurse-in-the-loop predictive control model for daily self-management of glucose and weight management in T2D patients. The online control model at its core utilizes an artificial neural network (ANN) predictive digital twin model that has been trained using transfer learning strategies to improve its performance (for both data rich and data scares setup) to generate food intake suggestions. As a proof of concept, we have the following contributions in this work-

1. A nurse-in-the-loop artificial intelligence feedback system.
2. A transfer-learned predictive digital twin model of patients.
3. An evolutionary optimization-based online control model in mHealth devices.

Figure *1* shows the overall schema of the proposed model. We first build a predictive model where inputs are the daily lifestyle choices, and the outputs are the next day's glucose and weight levels. Next, we used the predictive model as the core block for the online control model. The control model is built based on the Particle Swarm Optimization (PSO) method, which generates feedback that minimizes glucose and weight outputs. This is to be mentioned to cope with and further fine-tune the prediction model towards a patient; we re-trained the models every week while the PSO algorithm doesn't need further training. Instead, PSO utilizes the fine-tuned model as a core plant to generate diet suggestions.

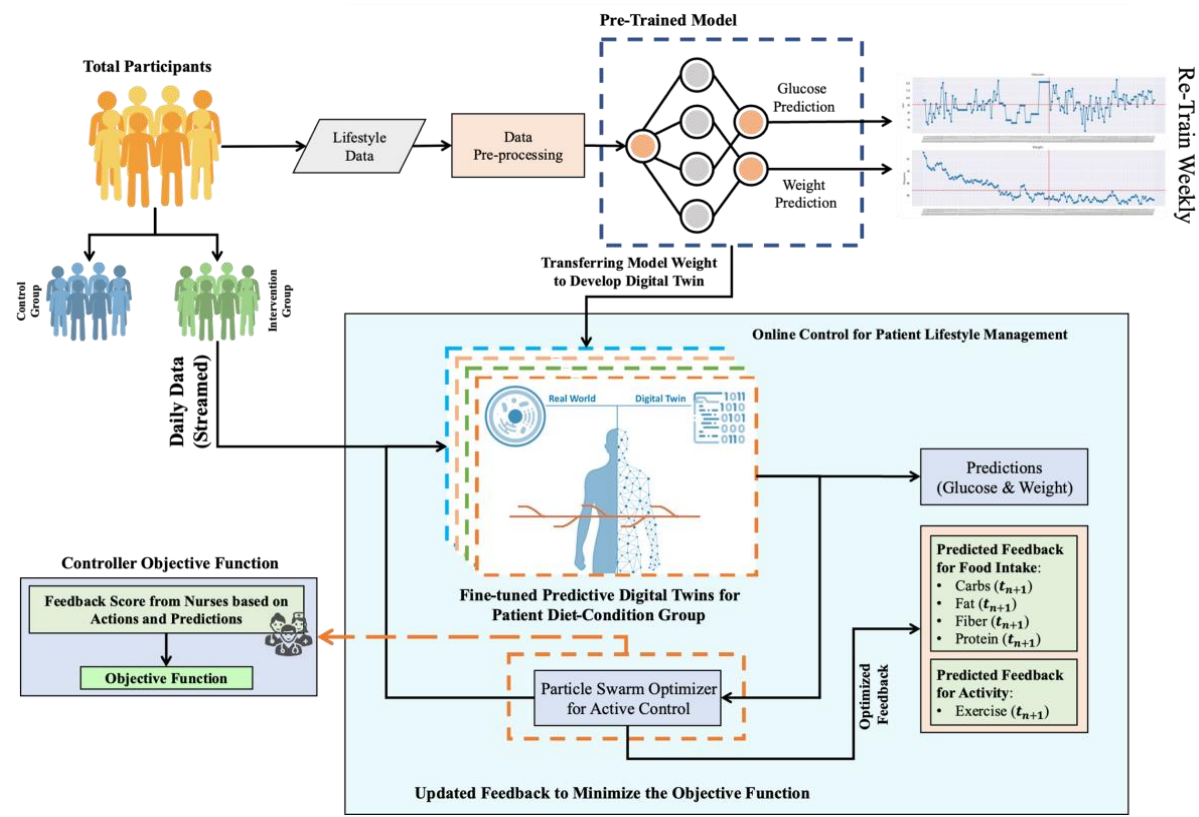

*Figure 1: Overall Scheme of the proposed method.*

## Methods:

This is an ancillary randomized controlled trial (RCT) that was deployed at the same time as the parent RCT. The model development and analysis consist of several steps, including data preprocessing, prediction model construction, optimization, evaluation, and finally, development of the online control model.

## Data Collection:

This study utilizes data collected from mobile health devices in a single center, 6-month, stratified, randomized controlled trial [27]. The parent trial recruited 60 overweight/obese adults (18+ years old), including 40 T2D patients. The study patients were equally randomized to a ketogenic or low-fat, low-calorie diet. The patients were to self-monitor their diet intake using a smartphone-based application. *Fitbit Inspire 2* was used to monitor daily physical activity and diet, and the *Withing's* body scale was used for weight self-monitoring. Additionally, the patients were provided health education based on evidence-based behavioral lifestyle interventions. Further details on the study can be found in Du et al [27].

This ancillary study used data from 20 T2D patients to develop the online control model. This selected sample was then randomly assigned into two groups: Group 1-Daily individualized feedback messages (Intervention group), Group 2- No feedback messages were provided (Control group). The data collected during the first three months of the study was used to train the predictive digital twin model. Continuous data from 3 months on were streamed directly to the online control model to generate daily suggestions and were sent to the participating subjects via text message starting from the end of 3 months. The predictive digital twin was also re-trained at the end of every week to improve the models' performance.

## Model Construction

### Predictive Digital Twin: Transfer Learning-Based Artificial Neural Network to Predict Patient States

Before developing the control model, we start by developing a predictive twin that will emulate the patient's behavior. By definition, a predictive twin models the future behavior and state of any system in consideration. These models utilize historical data to incorporate possible system modes within the model. We construct the digital twin to receive input (self-monitoring data) from patients (their food intake, weight, exercise regimen, etc.). In our case, these data are collected through mobile devices. This allows the twin to simulate the physical object in real-time, in the process offering insights into their performance (weight and blood glucose control).

We construct a simple multilayer perceptron/artificial neural network (with three hidden layers) (ANN) for predictive modeling of daily glucose levels using the health time series data. As the data to train the model contains missing/imputed data and based on the patients' physical conditions (and different diets) they experience a vast number of complex patterns, we utilize a transfer learning strategy. In literature, transfer learning strategies have proved to make learning from data efficient, fast, and under sparse data fruitful [15]. To achieve this goal, we concatenated and sampled all the patient's data and created the transfer learning dataset to pre-train the ANN model. Once the pre-training is done, the model is fine-tuned for each patient diet-condition group (*1. Keto Diet – Obese and T2D, 2. Keto Diet – Obese, Kidney Disease, and T2D, 3. Low-Fat Diet – Obese and T2D, 4. Low-Fat Diet – Obese, Kidney Disease, and T2D*). The training procedure involves sampling data to train a deep learning model for each patient group of interest, where the ANN model weights of the pre-trained model will be used as the prior. In the results section, we compare the performance of the model with classic machine learning models (*KNN, Gaussian Process, Decision Tree, Random Forest, Gradient Boosting*). Parameters for all the models are tuned using a grid search approach. While many of the above-mentioned algorithms can be used for building the predictive twin (comparable accuracy; later shown in the results section), we selected the transfer learning (TF)-based ANN algorithm for our analysis for two specific reasons:

(1) They adapt to learning complex patterns present in the data, and
(2) The predictive twin can be trained even if there is a low amount of data (data sparsity) for a new patient group.

Figure 2 shows a visual representation of the transfer learning strategy used in this work. We followed the same steps as Faruqui et [9]. The only change was that they fine-tuned their models for each patient, and we fine-tuned for each patient's diet-condition group.

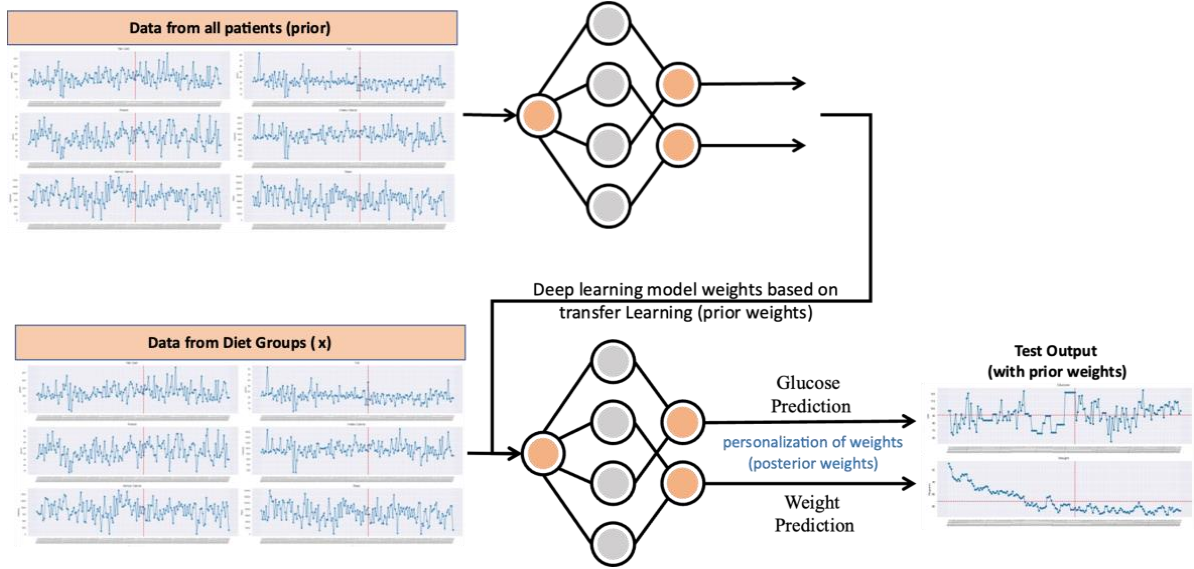

*Figure 2: Scheme of the transfer learning strategy to train an artificial neural network for predictive modeling. For patients with fewer observations, we pre-train a model with observations sampled from all the patients in the dataset. The pre-trained model is then fine-tuned with the data of the patient of interest.*

**Online Control Model: Particle Swarm Algorithm**

Once the predictive twin is developed, we focus on developing the control module of the proposed algorithm. The purpose of the control model is to use the predictive twin (of the patient-diet group) to determine what combination of food intake and exercise would help a patient maintain their blood glucose level and help them lose weight while maintaining their blood glucose level. This highly non-linear problem needs to be optimized efficiently within the search space. We deploy the Particle Swarm Optimization (PSO) [16] algorithm as the controller due to its simple implementation; it doesn't need a gradient/differential form of the objective function and, depending on the setting, is less memory consuming (can be used in both online and off-line setup). For a function defined in multidimensional vector space (as in our study case), PSO is a robust meta-heuristic optimization algorithm and one of the best optimization algorithms to find the minimum or maximum of the objective function. The PSO algorithm (Step 2 of *Figure 1*) is then used with the predictive twin to determine the best parameters to minimize the objective function. The PSO algorithm consists of two components: (1) the Objective function and (2) The constraints.

*The Objective Function*

To set up PSO for our case study, we need an objective function that helps us achieve our goal. For a patient in the low-fat diet group, our goal is to suggest a food and exercise regimen that can help them lose weight ($W$) till they reach their target weight level ($W_G = W_0 - 0.2W_0$). Once they reach their target level, we switch the goal to maintaining that weight. The patients will also need to maintain their Glucose levels ($G$) within an acceptable range ($70 \leq G \leq 130$). A patient in the keto diet group must attain (or close to) a certain blood ketosis level. To do that, They need to reach a keto ratio ($K$) of 1.5 or higher. The keto ratio can be computed from the suggested food intake (macronutrients) by the PSO algorithm. The objective function is special in the sense that we don't want the algorithm to select extreme parameters (in our case, food intake and physical activity suggestions) that cause the glucose level to go near zero or too much of a higher ketone level. To resolve this issue, we introduced a tuning parameter $\lambda$, which takes the value of 0 whenever a generated suggestion reaches its desired value;

otherwise, it takes a value of 1. This helped control the model to achieve its desired output. We integrated the feedback of the nursing knowledge using a second multiplier (penalty factor, $m$) for each element of the objective function. Depending on the suggestion provided by the controller, if the predicted glucose/weight/ketone level is too diverse, a nurse can assign a higher penalty and vice versa. In our case, this assigned score ranged from 1 to 1000 to penalize the model if it's too far away from desired values. At the end of each evaluation phase, the score derived from nursing knowledge is assigned to the model. The details of the decision boundaries considered by the nurses are provided in Appendix A.1. Thus, considering the above, the objective function for the PSO algorithm is set as below:

Objective Function

$$\min_{l=1,\ldots,L} \sum_{i=1}^{n} \lambda_1 m_1 \hat{G}_i + \lambda_2 m_2 \widehat{W}_i + \lambda_3 m_3 (1.5 - K_i)$$

Were,

Tuning parameter,

$\lambda_1 = 0$; if $70 \leq \hat{G}_i \leq 130$

$\lambda_2 = 0$; if $(W_i - \widehat{W}_i) \geq 0$ or $W_G = \widehat{W}_i$

$\lambda_3 = 0$; if *Patient-Diet Group* = 'keto-diet' or $(1.5 - K_i) \geq 0$

$m_{i=\{1,2,3\}}$ = Score assigned by a nurse-in-the-loop. This is used in conjuncture with the tuning parameters ($\lambda_{i=\{1,2,3\}}$)

$K = \frac{f_a}{(c - f_b + p)}$; $K$ = Keto ratio, $c$ = carb, $f_a$ = fat, $f_b$ = fiber, $p$ = protein

$W_G = W_0 - 0.2W_0$; $W_G$ = Weight Goal, $W_0$ = Patients weight at the start of the study

The scores ($m$) assigned by the nurse/s in the loop can also be automated future studies by:

1. Building a lookup table generated by the nurse/s for all the factors to be used during the optimization session. This will use ranges of values to assign penalty factors (for example, for glucose levels between 131 – 140, the penalty assigned can be 10).
2. Building an approximation linear function by using the lookup table. That way, we can approximate the penalty factors even if there is a value present that wasn't assigned a penalty by the nurse/s.

*The Constraints*

Furthermore, we also define the search space for the proposed model. The model constraints are decided based on the appropriate diet requirement [17,18] and discussions with nurse/s in the loop (nutritionists and nurses). The constraints may also be adapted (*exception case*) based on patient specific needs based on their physical condition and nutritional needs. Thus making the model adaptive not only patient group specific but also patient-specific. For this work, the general constraints were defined as follows-

|       Low-Fat Group       |        Keto Group        |
| :-----------------------: | :----------------------: |
| $195 \leq c \leq 300$     | $20 \leq c \leq 50$      |
| $20 \leq f_a \leq 55$     | $90 \leq f_a \leq 200$   |
| $20 \leq f_i \leq 50$     | $20 \leq f_i \leq 50$    |
| $100 \leq p \leq 160$     | $30 \leq p \leq 110$     |

## Results and Discussion

### Evaluation of Predictive Twin

We considered two variations of the proposed predictive twin model. The two variations are (1) The base ANN (without transfer learning) and (2) ANN-TF (with transfer learning). Several baseline methods are used to compare the performance of the proposed model. All the models are tuned using a grid search approach.

To evaluate the performance of the proposed model, we trained it using the first three months of data, during which no intervention was performed in both groups. We used one month of test data immediately following the third month's intervention. Although not explicitly included in the reported evaluation (as shown in Table 1), the predictive twin utilized for the control model was re-trained at the end of each week as the patient's new data became available. We evaluated all the models using Clark Error Grid [19]. Table 1 summarizes the percentage of prediction that falls on Zone A of the Clark Error Grid for all comparing methods.

*Table 1: Prediction accuracy of the proposed neural network models along with other comparing methods based on Clark Error Grids zone A.*

| Model | Keto Diet Group | | Low-Fat Diet Group | |
| --- | --- | --- | --- | --- |
| | Obese + Diabetes | Obese + Kidney + Diabetes | Obese + Diabetes | Obese + Kidney + Diabetes |
| | Glucose Level | Glucose Level | Glucose Level | Glucose Level |
| **KNN Regression (Weight: Distance)** | 77.50% | 77.50% | 77.50% | 77.50% |
| **KNN Regression (Weight: Uniform)** | 77.78% | 77.78% | 77.78% | 77.78% |
| **Gaussian Process (Kernel: RBF)** | 76.94% | 76.94% | 76.94% | 76.94% |
| **Decision Tree** | 71.67% | 71.67% | 71.67% | 71.67% |
| **Random Forest** | 84.17% | 84.17% | 84.17% | 84.17% |
| **Gradient Boosting** | 84.44% | 84.44% | 84.44% | 84.44% |
| **Artificial Neural Network (ANN)** | 79.72% | 87.44% | 87.78% | 84.23% |
| **ANN (with Transfer Learning)** | 80.00% | 87.85% | 87.22% | 84.90% |

The proposed model, ANN-TF, performs the best, followed by ANN, Gradient Boosting, and Random Forest. Both Random Forest and Gradient Descent showed significant performance (84.17% and 84.44%) in predicting the blood glucose level for patients in the Keto-Diet group with obesity and Diabetes. They are competitive with the proposed ANN-TF model for patients with a Low-Fat Diet group with obesity. Kidney disease and Diabetes (84.17% and 84.44%). Compared to these, the all-other method falls below 80% in performance, especially the Decision Tree, which has the lowest prediction performance (71.67%) across all the patient groups. Therefore, based on the evaluation in Table 1 (also considering the advantage of transfer learning whenever the data is sparse) and ease of use for control steps, we decided to utilize the ANN-TF model as the core/plant digital twin for the control model.

**Evaluation of the Control Model**

The objective of this study is to provide patients with dynamic suggestions (for food intake and exercise) every day to help them manage their blood glucose level and support a steady weight loss till they achieve their target weight. To achieve this, we deployed the PSO algorithm as a control model with a predictive twin. To evaluate the performance of the control model and its effect on patients, we separated the study patients into two sub-groups (as mentioned earlier). Group 1 started receiving feedback from the proposed model daily (intervention group), and Group 2 will not receive any feedback from the model (control model). The participants didn't receive any feedback for the first three months of the study. After the first in-person intervention (at month three), group 1 started receiving daily feedback messages. Group 2 continued to follow their suggested diet (the same diet they followed the first three months). Both groups of participants had the primary goal of –

1. Maintaining a steady weight loss till they achieve their weight goal was decided at the beginning of the study.
2. Maintaining their blood glucose level within a certain range.

This was an online setup, so we continued to receive patients self-monitoring data daily, and based on that, we continued to provide them with daily food and exercise feedback for three months.

**AI feedback on self-monitored physical activity and diet.**

Figure 3 shows the Self-monitored lifestyle data for the patients in this study. It shows the average monthly breakdown of the collected self-monitored data, including patients' average daily steps, daily caloric intake, total carb, and fat consumed across two distinct groups among the patients adhering to either a ketogenic diet or a low-fat diet. We identify the start of the message received by the red-dotted line in the figure. Over the six months, daily step counts were initially higher in the non-AI group as compared to the AI group. Interestingly, although a general decline in steps was noted across the six months for both groups, the AI group exhibited a rebound in physical activity starting at the four months, gradually approaching the step counts of the non-AI group while they showed a declining trend. The diet intake is visualized by diet groups as the diet regimen is different. Calorie intake in the KD group was stable and decreased in the AI group, while it increased in the non-AI group. In the LF group in which a low-calorie and low-fat diet was suggested, calorie intake decreased after AI feedback, while it increased in the non-AI group. For KD participants, the total carb intake maintained to be low in the AI group since the second month, while it increased in the LF group; for LF participants, carb intake increased around four months but started to decline in the AI group. In the KD group, despite the fat intake trending to decrease in both AI and non-AI groups, AI group participants, on average, maintained higher fat intake; fat intake in the LF group in the AI group maintained lower than in the non-AI group.

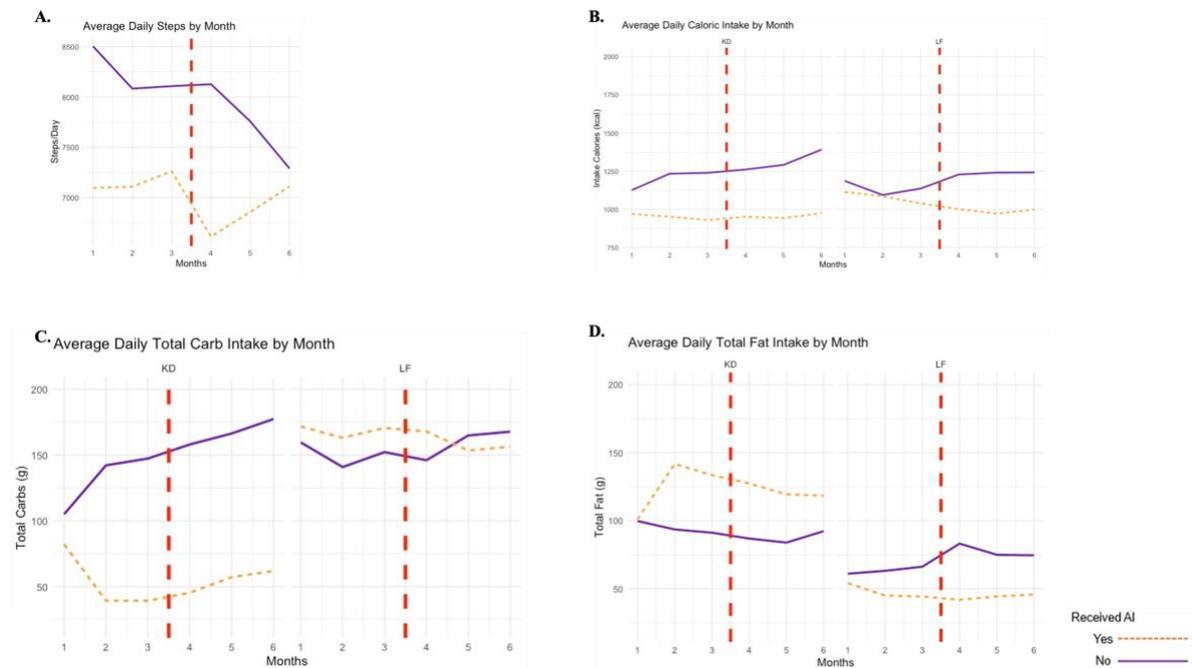

*Figure 3: Self-monitored lifestyles data of A. average daily steps by month; B. daily caloric intake by month between keto (KD) and low-fat (LF) group; C. daily total carb intake by month between KD and LF group; and D. daily total fat intake by month between KD and LF group. Physical activity was recommended the same for both groups, and KD was suggested to take low carb and high fat diet, while LF group was suggested to take low fat diet. Red dash vertical lines meant the time point (3-month) AI message started to be sent out.*

The AI group adhered to their weight loss target much better than the non-AI group. The non-AI group showed a smaller weight decrease (3.574 lbs) from a baseline mean of 92.29 lbs, while the control group exhibited a more considerable weight decrease (5.871 lbs) from a baseline mean of 97.35 lbs. Notably, the intervention effect was significant (time intervention $p = 4.731e-8$, group, $p = 0.310$, and interaction, $p = 0.423$). In terms of achieving their weight goals, four subjects in AI group (highest 14.82% weight loss) and two subjects in non AI group (highest 8.36% weight loss) reached their assigned goals. Both groups maintained glucose levels over time (p-value 0.661) compared to their respective baseline. In summary, we observed that AI-generated feedback may serve as a beneficial tool for participants to either sustain or recommit to their recommended lifestyle changes.

**Limitations**

There are several limitations to this study. As blood glucose and weight are not solely determined by lifestyle, other factors, such as age, gender, gut microbe, environment, and genes, also contribute to glucose and weight management. While our blood glucose prediction model is generalized for the study group, it may not be represented in the population. Also, considering the traits mentioned earlier are beyond the scope of this study, further reward systems could be developed in conjunction with the model to further motivate the patients to self-managing their daily food and exercise habits. In addition, other factors, such as food accessibility, which are important to help adherence to healthy lifestyles, were not evaluated in the current study. Lastly, the sample size was small with a relatively short follow-up period, future studies with a larger sample size and a longer study period are warranted.

# Conclusion

This study developed a novel nurse-in-the-loop predictive online control model that may promote recommended lifestyles for T2D self-management by providing AI-generated personalized feedback on daily lifestyle choices. The proposed model utilizes a transfer learning-based artificial neural network as a predictive digital twin to learn the characteristics of T2D patients following a specific diet and exercise regimen. Our findings revealed that participants who received AI-generated feedback maintained or reengaged their recommended physical activity and diet regimens. The model has the potential to be integrated into mobile health environments to support T2D individuals in managing their lifestyle behaviors for health benefits. Nonetheless, it should be acknowledged that our study was a proof-of-concept, and further research is warranted to optimize and validate the model. Future studies are recommended to utilize more rigorous study designs to examine the effect of the model on individuals' lifestyles and health outcomes. Furthermore, the generalizability of the model should be tested in a larger trial with longer follow-ups and a more diverse population to ensure its effectiveness in enhancing the weight and glycemic control of T2D patients.

# Acknowledgments

We would like to thank all study participants for their contributions to the study. Dr. Yan Du was partially supported by the RL5 Mentored Research Career Development Award (Trainee: Yan Du) through the San Antonio Claude D. Pepper Older Americans Independence Center (P30AG044271). Dr. Jing Wang was partially supported by the University of Florida–Florida State University Clinical and Translational Sciences Award funded by NIH UL1TR001427.

# Funding Source

The study was funded by the 80/20 Foundation (PI: Sharma). The funding source was not involved in the study design, data collection, data analysis, or writing of this manuscript.

# Conflict of Interest

The authors declare there is no Conflict of Interest.

# Data Sharing Statement

The data collected for this study, including individual participant data, will be made available upon request. The data will be shared in a deidentified format to ensure participant privacy and confidentiality. Access to the data will be granted following the approval of a detailed proposal and the signing of a data access agreement. This agreement will outline the terms of use, ensuring that the data are utilized for legitimate research purposes and in a manner that preserves the confidentiality and integrity of the data. Requests for data should be directed to Yan Du (duy@uthscsa.edu). The study protocol, statistical analysis plan, and informed consent form are accessible in the published protocol paper associated with this study.

# Appendix A:
## A.1. Model Scoring

The AI prediction ability was determined by the interventionists using a rating scale of

1. Bad (score: 1000),
2. Okay (score: 500),
3. Good (score: 100), and
4. Very Good (score: 1).

The important levels of A.I. predicted variables (e.g., intake calories, fat in grams, weight) were classified into 3 categories.

1. Very Important,
2. Moderately Important, and
3. Low Importance.

Figure A.1. below illustrates the workflow the nurse-in-the-loop used to assess and enhance the A.I. model's performance over time for the subjects in the study. The boundaries set for each subject is based on the in-person assessment done when the study started (the actual boundaries not shown in the appendix). The hard boundaries set for each variable had to be met depending on participant's diet group and personal characteristics (see Tables A. 1-2).

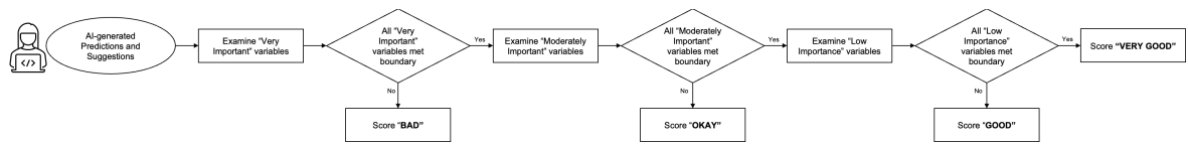

**Figure A. 1.** "Expert-in-the-loop" Workflow

**Table A. 1.** "Hard" boundaries for the ketogenic diet group.

| Variables | Importance Rank | Boundary |
| --- | --- | --- |
| Net Carb | Very important | 20-50 |
| Keto Ratio (calculated) | Very important | $\geq 1.5$ |
| Weight | Very Important | $\pm 5$ lbs. |
| Blood Glucose | Very important | 70-130 |
| Protein | Moderately Important | $\geq$ minimum protein |
| Fat | Moderately Important | $\geq$ minimum fat |
| Intake Calories (calculated) | Moderately Important | Lower than calorie goal |
| Blood Ketone | Moderately Important | $\geq 0.5$ |
| Activity Calories | Low Importance | $< 500$ kcal + intake calories |
| Steps | Low Importance | $\geq 6000$ |

**Table A.2.** "Hard" boundaries for the low-fat diet group.

| Variables | Importance Rank | Boundary |
|---|---|---|
| Carb | Moderately Important | < 65% calories from carb |
| Protein | Moderately Important | ≥ minimum protein |
| Fat | Very Important | < maximum fat |
| Intake Calories | Very Important | Lower than calorie goal |
| Activity Calories | Low Importance | < 500 kcal + intake calories |
| Steps | Low Importance | ≥ 6000 |
| Weight | Very Important | ± 5 lbs. |
| Blood Glucose | Very Important | 70-130 |

**A.2. Translate A.I. Predictions and Suggestions to Deliverable Messages**

There were three components of daily text messages (1) a meal plan example, (2) tailored motivational messages, and (3) a daily step goal.

The meal plan example was created using the linear programming approach. We generated a list of five food groups - meat, fruits, vegetables, nuts and seeds, and saturated fats - with food items available from the ADA website. The objective of the linear programming approach was to ensure the variety of food items while meeting the daily nutrient needs and daily caloric intake limits for ketogenic diet and low-fat diet participants. Specifically, each individual had unique constraints considered while building the linear programming model. The linear programing algorithm selected the food items and serving sizes from different food groups to create a personalized meal plan (Figure A.2. shows a sample meal plan that is provided to the subjects of the intervention group).

| Food Group | Meal Plan Example |
|---|---|
| Lean-Fat Meat and Substitutes | 4 servings of (1/4 cup) Any cottage cheese |
| Medium-Fat Meat and Substitutes | 3 servings of (1 oz) Mozzarella |
| High-Fat Meat and Substitutes | 3 servings of (1 oz) Spareribs, ground pork, pork sausage (patty or link) |
| Vegetables | 2 servings of (1/2 cup) Mushrooms, cooked |
| Fruits | 2 servings of (3/4 cup) Blueberries (raw) |
| Whole Milk | 1-2 cups of milk or milk substitutes of your choice |
| Nuts and Seeds | 5 servings of (1 tsp) Oil (corn, cottonseed, safflower, soybean, sunflower, olive, peanut) |
| Saturated Fats | 5 servings of (2 tbsp) Cream (light) |

**Figure A.2.** Daily Message: Meal Plan Example

Prior to the initiation of the intervention, the interventionist generated a motivational message pool consisting of 200+ messages that covered various domains including "Positive Feedback", "Carbohydrate", "Protein", "Fat", "Fiber", "Overall Nutrition", "Self-monitoring", and "Exercise". During the intervention, the interventionist evaluated areas need to be improved for each individual and then chose a message from the relevant domain to be sent.

To establish a daily step goal, we utilized the based on the 70[th] percentile rank of the last 10 days step data and sent together with an exercise motivational quote.

**Table A.3.** Example of AI suggestions for one ketogenic diet group participant

|  | Net Carb | fat | protein | intake calories | activity calories | steps | glucose | ketone | weight |
|---|---|---|---|---|---|---|---|---|---|
| Last observation | 39 | 45.2 | 104.1 | 1064 | 1009 | 5253 | 134 | 0.2 | 199.2 |
| AI Suggestion | 30 | 135 | 60 | 1064 | 1008 | 6000 |  |  |  |
| Predicted Outcome |  |  |  |  |  |  | 110 | 2.4 | 197.6 |
| Keto Ratio (Last Observation) | 0.3 (vs. 1.5) | | | | | | | | |
| Keto Ratio (AI Suggestion) | 1.2 (vs. 1.5) | | | | | | | | |